\documentclass[10pt,twocolumn,letterpaper]{article}

\usepackage[pagenumbers]{cvpr}

\usepackage{graphicx}
\usepackage{amsmath}
\usepackage{amssymb}
\usepackage{booktabs}
\usepackage{url}

\usepackage{utfsym}
\usepackage{pifont}
\usepackage{soul}
\usepackage{dsfont}

\usepackage{subcaption}
\usepackage{caption}

\usepackage[pagebackref,breaklinks,colorlinks]{hyperref}

\usepackage[capitalize]{cleveref}
\crefname{section}{Sec.}{Secs.}
\Crefname{section}{Section}{Sections}
\Crefname{table}{Table}{Tables}
\crefname{table}{Tab.}{Tabs.}

\makeatletter
\newcommand{\thickhline}{%
  \noalign {\ifnum 0=`}\fi \hrule height 1pt
  \futurelet \reserved@a \@xhline
}
\makeatother

\def\cvprPaperID{*****}
\def\confName{CVPR}
\def\confYear{2025}

\begin{document}

\title{A DVDrive Approach for doScenes Instructed Driving Challenge}

\author{
Team MIC Lab \\
Zijian Fu $^{1}$, Xiangyang Chu$^{2}$, 
Mengshi Qi$^{1}$\thanks{Corresponding author}, Huadong Ma$^{1}$, Guanghao Zhang$^{2}$, and Wei Li$^{2}$\thanks{Project leader. Work done while Zijian Fu was an intern at Xiaomi EV.}\\
$^{1}$Beijing University of Posts and Telecommunications, $^{2}$ Xiaomi EV\\
{\tt\small \{zijianfu, qms, mhd\}@bupt.edu.cn}
}

\maketitle

\begin{abstract}
Instruction-conditioned trajectory prediction is an emerging problem in autonomous driving, where a model predicts the future ego trajectory not only from visual scene context and historical motion, but also from a natural-language maneuver instruction. This paper presents our submission to the doScenes Instructed Driving Challenge, built upon OmniDrive, a vision-language-action driving agent with 3D perception, reasoning, and planning capabilities. We adapt OmniDrive to the doScenes setting by training it on instruction-annotated nuScenes scenes and generating a 6-second ego trajectory represented by 12 future waypoints. To improve multi-view visual grounding, we further introduce a DVPE-style divided-view perception module into the OmniDrive perception head. Instead of attending globally to all camera features, the proposed module groups query features and image tokens into divided local view spaces and performs visibility-aware cross-attention within each view. This design reduces irrelevant cross-view interference and helps the model better align language instructions with local driving-relevant visual evidence. The code is publicly available at:
\url{https://github.com/feel12348/doscenes-omnidrive}.

\end{abstract}

\section{Introduction}
\label{sec:intro}
Trajectory prediction is a fundamental component of autonomous driving. Given historical ego motion and surrounding scene observations, the model is expected to predict a future trajectory that is both geometrically plausible and consistent with the driving context. Most existing trajectory prediction benchmarks rely mainly on past motion, perception features, and map information. However, for ego-vehicle planning, the future intention is often available before the maneuver is executed. For example, a human passenger or navigation system may provide an instruction such as “turn left at the next intersection” or “change to the right lane.” Such instructions can reduce ambiguity in future motion prediction.
The doScenes~\cite{roy2025doscenes} Instructed Driving Challenge formulates this problem as instruction-conditioned ego-trajectory prediction.

This setting is naturally suited for vision-language-action models, as the model is required to jointly interpret visual scenes, follow language instructions, and generate spatially grounded trajectory outputs. OmniDrive~\cite{wang2025omnidrive} provides a strong foundation for this task by introducing a holistic driving vision-language framework that integrates 3D perception, reasoning, and planning. It employs sparse queries to lift and compress multi-view visual representations into a compact 3D-aware feature space. However, its global multi-view interaction may still introduce redundant features from spatially irrelevant regions, which can weaken local perception and increase the difficulty of long-horizon trajectory generation. To address this limitation, we further draw inspiration from DVPE~\cite{wang2024dvpe}, which improves multi-view 3D perception by decomposing the global space into divided local views and performing position-aware feature interaction within each view. Building on this idea, our method enhances OmniDrive with divided-view multi-view perception, enabling more effective local spatial reasoning for navigation-conditioned trajectory prediction.

\section{Related Work}
\label{sec:related}

\subsection{End-to-End Driving and Trajectory Prediction}
Trajectory prediction and motion planning are central to modern autonomous driving systems. Early end-to-end approaches such as ST-P3~\cite{hu2022stp3} learn spatial-temporal features to jointly perform perception, prediction, and planning from raw sensor inputs. UniAD~\cite{hu2023uniad} unifies detection, tracking, mapping, motion, and occupancy prediction into a single planning-oriented framework, while VAD~\cite{jiang2023vad} adopts a vectorized scene representation to improve planning efficiency. These methods condition future motion mainly on visual observations and historical states. In contrast, the doScenes~\cite{roy2025doscenes} setting, built upon nuScenes~\cite{caesar2020nuscenes}, additionally provides natural-language navigation instructions, which our method exploits to reduce intention ambiguity in long-horizon trajectory prediction. Beyond planning, many techniques have been explored to improve the robustness, reasoning, and safety of driving systems, such as test-time adaptation for robust detection~\cite{chen2024improvingbn}, knowledge-graph-based retrieval-augmented generation~\cite{qi2025safedriverag}, traffic topology scene-graph reasoning~\cite{qi2025t2sg}, and driver attention prediction~\cite{qi2023unsupervised}.

\subsection{Vision-Language Models for Driving}
Recent work integrates large vision-language models into driving to enable reasoning and language-level interaction. Talk2Car~\cite{deruyttere2019talk2car} is among the first to ground natural-language commands in driving scenes. GPT-Driver~\cite{mao2023gptdriver} and DriveGPT4~\cite{xu2024drivegpt4} reformulate planning as language generation, producing trajectories or control signals from textual prompts, while DriveLM~\cite{sima2024drivelm} introduces graph-structured visual question answering for driving, and LMDrive~\cite{shao2024lmdrive} performs closed-loop driving conditioned on language instructions. OmniDrive~\cite{wang2025omnidrive} provides a holistic vision-language-action framework with 3D perception, reasoning, and counterfactual planning, which we adopt as our baseline and extend with instruction-conditioned trajectory supervision. More broadly, our work connects to the rich literature on multi-modal video understanding, including in-context segmentation~\cite{qi2025dcsam}, audiovisual commonsense reasoning~\cite{qi2023robust}, action quality and form assessment~\cite{qi2024aqa, qi2025explainable}, multi-modal 3D human pose estimation~\cite{qi2025balanced}, temporal action localization~\cite{yun2024weakly}, cross-modal video retrieval~\cite{qi2021semantics}, few-shot video classification~\cite{qi2020fewshot}, and sports video captioning~\cite{qi2020sports}, all of which highlight the importance of effectively fusing visual and language cues.

\subsection{Multi-View 3D Perception}
Robust 3D perception from surround-view cameras is a prerequisite for reliable trajectory prediction. DETR3D~\cite{wang2022detr3d} introduces 3D-to-2D queries to aggregate multi-view features, and BEVFormer~\cite{li2022bevformer} builds bird's-eye-view representations via spatiotemporal attention. PETR~\cite{liu2022petr} encodes 3D position information into image features, and StreamPETR~\cite{wang2023streampetr} extends this line with object-centric temporal modeling. However, these methods typically perform global cross-attention over all views, which introduces redundant cross-view interactions. DVPE~\cite{wang2024dvpe} alleviates this issue by dividing the global space into local views with view-specific position embeddings. Inspired by DVPE, our DVDrive module performs visibility-aware divided-view perception within OmniDrive to strengthen local spatial reasoning for instruction-conditioned trajectory prediction.

\section{Method}

\subsection{Problem Formulation}

We study the instruction-conditioned ego-trajectory prediction task in doScenes. 
Given multi-view camera observations, historical ego motion, map-related information, and a natural-language driving instruction at the current timestamp $t$, the model is required to predict the future ego trajectory over the next $6$ seconds. Following the official doScenes evaluation protocol, the future trajectory is sampled at $2$ Hz and therefore contains $12$ two-dimensional waypoints.

\subsection{Task Adaptation from OmniDrive to doScenes}
We adopt OmniDrive as the baseline framework and adapt it to the doScenes trajectory prediction task. Although OmniDrive provides a strong vision-language-action foundation for autonomous driving, its original training setting is not directly aligned with the requirements of doScenes. Specifically, the original OmniDrive training pipeline does not explicitly incorporate navigation instructions as conditional inputs for trajectory generation. As a result, the model has limited ability to adjust future motion planning according to language-level navigation intentions. In addition, OmniDrive is mainly trained with short-horizon trajectory supervision, where the future trajectory is represented within a 3-second prediction window. In contrast, doScenes requires the model to predict a 6-second future ego trajectory, which introduces a longer temporal reasoning requirement.

To bridge this gap, we first reconstruct the training data according to the doScenes evaluation protocol. For each training sample, we extract the future ego trajectory from the corresponding driving scene and extend the supervision horizon from 3 seconds to 6 seconds. This data adaptation enables the model to learn longer-term motion evolution and generate trajectories that are consistent with the target benchmark. Furthermore, during supervised fine-tuning, we randomly sample one navigation instruction associated with the current scene and append it to the textual prompt. In this way, the model is optimized under both visual scene observations and language-based navigation intentions, allowing it to produce future trajectories that are better aligned with the driving context and the given instruction.

\subsection{DVDrive: Divided-View Multi-View Perception}

After adapting OmniDrive to the doScenes trajectory prediction setting, and inspired by DVPE~\cite{wang2024dvpe}, we further introduce a divided-view multi-view perception module, termed DVDrive, to enhance its spatial understanding of 3D driving scenes. The original OmniDrive relies on global multi-view feature interaction to connect visual observations with the language-action decoder. However, such a global interaction requires each query to attend to image tokens from all camera views and all spatial regions. Many of these tokens are weakly related to the current local driving context, which may introduce redundant visual interference and increase the optimization difficulty of cross-attention. This problem becomes more evident in long-horizon trajectory prediction, where the model needs to capture fine-grained local structures such as lane boundaries, nearby agents, and drivable areas.

To address this issue, DVDrive follows the divided-view principle and decomposes the global 3D perception space into multiple local virtual views. Instead of performing cross-attention over the entire multi-view feature space, queries and image tokens are assigned to local views according to their 3D spatial locations. Each query mainly interacts with visual tokens located in the same local view, allowing the model to focus on geometrically relevant and locally visible image regions. This design reduces interference from unrelated camera views and distant spatial regions while preserving the ability to aggregate multi-view information.

Given multi-view image features from surrounding cameras,
\begin{equation}
    \mathcal{F} = \{F_c\}_{c=1}^{C},
\end{equation}
where $C$ denotes the number of cameras, we first project them into a unified Transformer feature space:
\begin{equation}
    \mathcal{M}
    =
    \mathrm{Proj}
    \left(
    \mathrm{Flatten}(\mathcal{F})
    \right)
    \in \mathbb{R}^{L \times d},
\end{equation}
where $L=C \times H \times W$ is the total number of multi-view image tokens and $d$ is the hidden dimension.

For each image token, we construct a 3D geometric representation by sampling multiple depth points along its corresponding camera ray and projecting these points into the ego-centric coordinate system:
\begin{equation}
    p_{j,d}^{3D}
    =
    \Pi_c^{-1}(u_j, v_j, z_d),
    \quad d=1,\ldots,D,
\end{equation}
where $(u_j,v_j)$ denotes the image location of the $j$-th token, $z_d$ is the sampled depth value, and $\Pi_c^{-1}(\cdot)$ denotes the inverse projection from the image plane to the ego-centric 3D space. Compared with conventional 2D image positions, such ray-based 3D coordinates provide richer geometric cues for multi-view spatial reasoning.

The global 3D perception space is then divided into multiple local virtual views. For the $s$-th shifted state and the $v$-th local view, the query set and visual token set are defined as
\begin{equation}
    \mathcal{Q}_{v}^{s}
    =
    \{q_i \mid r_i \in \Omega_{v}^{s}\},
    \quad
    \mathcal{K}_{v}^{s}
    =
    \{m_j \mid \bar{p}_{j}^{3D} \in \Omega_{v}^{s}\},
\end{equation}
where $r_i$ is the 3D reference point of query $q_i$, $m_j$ is the $j$-th visual token, $\bar{p}_{j}^{3D}$ is the representative 3D position of the visual token, and $\Omega_{v}^{s}$ denotes the corresponding local virtual view. In this way, each query mainly interacts with visual features within the same local view, rather than attending to all tokens in the global space.

For each local virtual view, both query reference points and visual-token coordinates are transformed into a view-specific coordinate system:
\begin{equation}
    \tilde{r}_{i}^{v,s}
    =
    \mathcal{T}_{v}^{s}(r_i),
    \quad
    \tilde{p}_{j,d}^{v,s}
    =
    \mathcal{T}_{v}^{s}(p_{j,d}^{3D}),
\end{equation}
where $\mathcal{T}_{v}^{s}(\cdot)$ denotes the coordinate transformation from the global ego-centric space to the local virtual view. Position embeddings are then generated based on the transformed local coordinates:
\begin{equation}
    e_{i}^{q,v,s}
    =
    \phi_q
    \left(
    \mathrm{PE}(\tilde{r}_{i}^{v,s})
    \right),
\end{equation}
\begin{equation}
    e_{j}^{k,v,s}
    =
    \phi_k
    \left(
    \mathrm{PE}(\tilde{p}_{j,1:D}^{v,s})
    \right),
\end{equation}
where $\phi_q$ and $\phi_k$ denote the query position encoder and the visual-token position encoder, respectively. By generating position embeddings in local coordinate systems, the model avoids learning complex spatial relations directly in the entire global coordinate space. This helps reduce the difficulty caused by large viewpoint variations across cameras.

Within each divided local view, DVDrive performs visibility-aware local cross-attention between queries and valid visual tokens:
\begin{equation}
    h_i^{v,s}
    =
    \mathrm{softmax}
    \left(
    \frac{
    (q_i + e_i^{q,v,s})
    (K_v^s + E_v^{k,v,s})^\top
    }{
    \sqrt{d}
    }
    +
    A_v^s
    \right)
    V_v^s ,
\end{equation}
where $K_v^s$ and $V_v^s$ denote the key and value features within the current local view, $E_v^{k,v,s}$ denotes the corresponding local position embedding, and $A_v^s$ denotes the valid-region attention mask. The attention operation is restricted to the effective feature set inside the current local view, so each query aggregates information mainly from spatially relevant regions. Compared with global cross-attention, this local interaction reduces interference from irrelevant visual tokens and improves the stability of multi-view feature aggregation.

In addition, DVDrive retains temporal memory modeling in the perception branch, allowing current-frame features to be aligned and fused with historical scene information. Such temporal context is important for long-horizon trajectory prediction, because future ego motion depends not only on the current observation but also on the evolution of surrounding agents and road structures over time.

\section{Experiments}
\label{sec:exp}

\subsection{Implementation Details}

We implement our method based on the OmniDrive framework and train it under the doScenes 6-second trajectory prediction setting. The model is trained with 12 future waypoints, corresponding to a 6-second prediction horizon. During supervised fine-tuning, doScenes navigation instructions are enabled, and one instruction associated with the current scene is randomly sampled as the language condition. We train the model using AdamW with an initial learning rate of $1\times10^{-4}$ and a weight decay of $1\times10^{-4}$. A cosine annealing learning rate schedule is applied, with 500 warm-up iterations and a minimum learning rate ratio of $10^{-3}$. The model is optimized for 6 epochs using an iteration-based runner. The batch size is set to 2 per GPU. All experiments are conducted on 8 NVIDIA RTX A6000 GPUs. The full training process takes approximately 24 hours.

\subsection{Experimental Results}

Since the labels of the official test set are not accessible, we report the evaluation results on the validation set. Following the doScenes evaluation protocol, we use ADE@1s, ADE@2s, ADE@3s, and mADE as the main metrics. Here, mADE denotes the average ADE over the evaluated future timestamps. Lower values indicate better trajectory prediction performance.

\begin{table}[t]
\centering
\caption{Validation results on doScenes.}
\label{tab:val_results}
\scriptsize
\setlength{\tabcolsep}{3pt}
\begin{tabular}{lcccc}
\toprule
Method & A@1s & A@2s & A@3s & mADE \\
\midrule
OmniDrive w/ Nav. & 0.151 & 0.328 & 0.637 & 0.372 \\
Ours w/ Nav.      & \textbf{0.145} & \textbf{0.316} & \textbf{0.613} & \textbf{0.358} \\
\bottomrule
\end{tabular}
\end{table}

As shown in Table~\ref{tab:val_results}, our method consistently improves over the navigation-conditioned OmniDrive baseline across all evaluation metrics. Specifically, our model reduces mADE from 0.372 to 0.358, demonstrating the effectiveness of incorporating navigation-conditioned training and DVPE-style divided-view perception. The consistent improvements at 1s, 2s, and 3s indicate that the proposed perception enhancement benefits both short-term and longer-horizon trajectory prediction.

\subsection{Challenge Leaderboard}

We further report the official ranking of the doScenes Instructed Driving Challenge. As shown in Table~\ref{tab:leaderboard}, on the hidden test set of the Full Multimodal track, our submission from Team MIC Lab ranks \textbf{first} among all participating teams, achieving an ADE of $2.0913$ and an FDE of $5.5946$. Compared with the official paper baselines (an ADE of $7.1086$ for the history-only baseline and $7.2046$ for the instruction baseline), our method reduces the displacement error by a large margin. Moreover, our ADE is clearly lower than that of the second-place team ($2.6774$), confirming the strong generalization of the proposed navigation-conditioned training and divided-view perception design on unseen scenes.

\begin{table}[t]
\centering
\caption{Official leaderboard of the doScenes Instructed Driving Challenge (Full Multimodal track, hidden test set). Our team, MIC Lab, ranks first. Best results are in \textbf{bold}; ``--'' denotes entries not provided on the leaderboard.}
\label{tab:leaderboard}
\scriptsize
\setlength{\tabcolsep}{5pt}
\begin{tabular}{clcc}
\toprule
Rank & Team / Model & ADE $\downarrow$ & FDE $\downarrow$ \\
\midrule
-- & History-Only Baseline & 7.1086 & 13.8567 \\
-- & Instruction Baseline  & 7.2046 & 13.9224 \\
\midrule
1 & \textbf{MIC Lab (Ours)} & \textbf{2.0913} & 5.5946 \\
2 & NudgeVAD                & 2.6774 & \textbf{5.5673} \\
3 & UCF UrbanITY Lab        & 2.7577 & 6.3753 \\
4 & why                    & 3.3309 & 8.1090 \\
\bottomrule
\end{tabular}
\end{table}

{\small
\bibliographystyle{ieee_fullname}
\bibliography{egbib}
}

\end{document}